# Supervised Contrastive Learning for Multimodal Unreliable News Detection in COVID-19 Pandemic

Wenjia Zhang
University of Warwick
United Kingdom
Wenjia.Zhang@warwick.ac.uk

Lin Gui
University of Warwick
United Kingdom
Lin.Gui@warwick.ac.uk

Yulan He
University of Warwick
Alan Turing Institute
United Kingdom
Yulan.He@warwick.ac.uk

## ABSTRACT

As the digital news industry becomes the main channel of information dissemination, the adverse impact of fake news is explosively magnified. The credibility of a news report should not be considered in isolation. Rather, previously published news articles on the similar event could be used to assess the credibility of a news report. Inspired by this, we propose a BERT-based multimodal unreliable news detection framework, which captures both textual and visual information from unreliable articles utilising the contrastive learning strategy. The contrastive learner interacts with the unreliable news classifier to push similar credible news (or similar unreliable news) closer while moving news articles with similar content but opposite credibility labels away from each other in the multimodal embedding space. Experimental results on a COVID-19 related dataset, ReCOVery, show that our model outperforms a number of competitive baseline in unreliable news detection.[1]

**ACM Reference Format:**

## 1 INTRODUCTION

Recent years have witnessed a growing proliferation of online sites that serve as channels for instantaneous information dissemination. However, they can also be used for spreading hoaxes and false information. In March 2020, the COVID-19 has been declared a pandemic by the World Health Organisation (WHO), which creates a breeding ground for fake news. Misinformation about diagnostic tests and immunisation campaigns could create panic, fragment social response, and also put people's lives in danger.

Traditionally, news credibility has been assessed by identifying the authenticity of a given news article [29] through binary classification [19], clustering [17], or knowledge-based methods [23]. Approaches to fake news detection are largely text-based, including classifiers built on hand-craft features [18], and deep neural networks, such as recurrent neural networks [1, 13], convolutional neural networks [26], graph neural networks [7, 12, 16], and adversarial networks [14]. Recently, multimodal approaches attempt to learn the incongruity expressions between text and image modalities by using attention mechanism to capture the interaction between modalities [6, 20, 24, 27]. Existing methods however only focus on the correlation between texts and images within input articles, and ignore the subtle relationships across articles. In this paper, we propose a novel contrastive learning framework which makes use of the most similar credible/unreliable news published prior to a given article to refine the learned metrics in the multimodal embedding space to overcome the aforementioned issue.

More concretely, inspired by the Vision Transformer [3, 11], we propose a unified BERT-based learning framework which compares both textual and visual information in a latent space based on the multi-head attention mechanism to extract features more effectively from multiple modalities. A constrative learning strategy is developed to encourage similar credible news articles (or similar unreliable news) to be closer and similar but with opposite credibility class labels to be far away in the multimodal embedding space.

To train the model more effectively with contrastive learning, we propose the use of a memory bank by buffering learned representations of input articles in the current epoch for similarity calculation of news articles in the contrastive loss in the next epoch. We have evaluated the proposed method on a COVID-19 related dataset, ReCOVery[2], and the results show our model outperforms several strong baselines. Our main contributions are two-fold: (1) To the best of our knowledge, the proposed model represents the first attempt of employing contrastive learning for multimodal unreliable news detection; (2) Each input image is segmented into a sequence of image patches to construct a unified BERT-like multimodal model for a better capture of subtle interactions between text and image.

## 2 RELATED WORK

*Fake News Detection.* Early approaches to fake news detection are mainly based on text features and are built on various neural network architectures [1, 7, 12, 13, 16, 26]. Recently, Zhou et al. [28] proposed a multimodal fake news detection approach which leverages a pre-trained image captioning model [22] to first generate a text description for the image in a news article and then concatenate it with news text to train a text-based fake news classifier. However, the quality of text generated from the image captioning model limits the performance of fake news detection. In addition, such an approach cannot capture the subtle interaction between text and image features. Other multimodal approaches first map text and image into their respective embedding spaces and then fuse multimodal features through simple concatenation, using attention mechanisms or tensor networks [6, 20, 24, 27]. Existing methods however only focus on modelling the relationship between text and image within a single article, ignoring relationships across articles. We instead propose a novel contrastive learning framework which makes use of the most similar credible/unreliable news published prior to a given article to refine the learned metrics in the multi-modal embedding space to achieve better unreliable news detection performance.

---

[1] Our source code can be accessed at https://github.com/WenjiaZh/BTIC.

[2] https://github.com/apurvamulay/ReCOVery

*Contrastive Learning.* Contrastive learning aims at learning an embedding space where positive pairs are close to each other while negative ones are far apart. It is also closely related to metric learning which aims at learning a distance function in an embedding space [25]. Supervised contrastive learning methods [8] employ both supervised classification losses and contrastive loss. They have achieved many successes in tasks such as visual representation learning [4] and few-shot learning [15]. However, to the best of our knowledge, no study leverages supervised contrastive learning for multimodal fake news detection. One possible reason is that in typical supervised contrastive learning, it is assumed that the degree of irrelevance between positive samples and randomly selected negative samples can be captured by the contrastive loss by comparing their class labels. However, in fake news detection, articles may be irrelevant even when sharing the same label as they may discuss different events or topics. Therefore, we need to develop a different strategy in choosing negative samples for contrastive learning in unreliable news detection.

## 3 METHODOLOGY

We propose a multimodel unreliable news detection model consisting of three main components: the text encoder, the image encoder, and joint learning with the integration of both cross-entropy loss and constrastive loss with the use of a memory bank. An overview of the mode is illustrated in Figure 1.

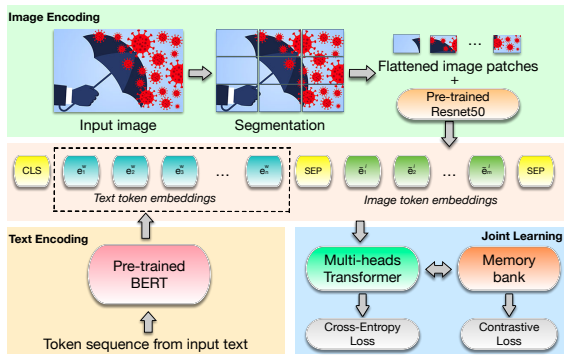

Figure 1: The framework of our proposed approach.

### 3.1 Text Encoder

The input to the text encoding component is a fixed-length word sequence from a news article $w = \{w_1, w_2, ..., w_n\}$, $n$ is the sequence length. A pre-trained BERT model [2] is used to encode the text sequence into a sequence of embeddings, $e^w = \{e_1^w, e_2^w, ..., e_n^w\}$ where $e^w \in \mathbb{R}^{n \times d}$, $d$ is the dimension of the word embedding.

### 3.2 Image Encoder

Inspired by the Vision Transformer [3, 11], we resize and split each image $I$ into $m$ small patches, $\{I_1, I_2, ..., I_m\}$, $I_j \in \mathbb{R}^{H \times W \times C}$, where $H$ and $W$ denotes the height and the width of an image patch, and $C$ is the number of channels used to represent the image patch. We set the resolution of the resized input image to $(H \times W) = (640 \times 640)$. Each image patch is encoded into a vector $e_j^i \in \mathbb{R}^{d^I}$ using a pre-trained ResNet-50 [5]. Here, $d^I$ is the dimension of the image patch representation. An image can then be represented as a sequence of constituent image patch vectors, $e^i = e_1^i \oplus e_2^i \oplus ... \oplus e_m^i$, $e^i \in \mathbb{R}^{m \times d^I}$, where $\oplus$ is the concatenation operator. In order to extract more informative features, its representation $e^i$ first goes through two linear projections to map from $d^I$-dimension to $d$-dimension. It is then fed to a Transformer layer to capture the subtle interplay between image patches. The final image representation is given by:

$$\tilde{e}^i = \text{Transformer}(W_2(W_1 e^i + b_1) + b_2) \quad (1)$$

where $W_1, W_2, b_1, b_2$ are trainable parameters to be learned in the linear projections.

### 3.3 Joint Learning

*Multimodal Feature Representation.* To integrate the extracted textual with visual information, we first concatenate the aforementioned representations as: $e = e^w \oplus \tilde{e}^i, e \in \mathbb{R}^{l \times d}$, where $l = n + m$ is fixed to the sum of the text length and the number of image patches with zero padding, the special [SEP] and [CLS] tokens. We then employ a multi-head self-attention mechanism [21] to relate the words with image patches. The final representation $x$ is given by:

$$x = \text{Concat}(head_1, ..., head_R)W^o$$
$$head_i = \text{Softmax}\left(\frac{[W_i^Q e]^\intercal [W_i^K e]}{\sqrt{d_k}}\right)[W_i^V e]^\intercal \quad (2)$$

where $R$ is the number of attention heads, $d_k = d/h$, $W_i^Q$, $W_i^K$, $W_i^V \in \mathbb{R}^{d \times d_k}$ are trainable parameter matrices.

*Cross-Entropy Loss.* After obtaining the multimodal features representation $x$, we use max pooling and average pooling to perform down-sampling, followed by a dropout with a probability of $p = 0.5$. The resulting final article representation $x_d$ is fed to a sigmoid layer to generate the predicted class label: $\hat{y} = \text{Sigmoid}(W_c x_d + b_c)$, where $W_c$ and $b_c$ are parameters to be learned. The cross-entropy loss is defined as: $\mathcal{L}_c = -\sum_{c=1}^{C} y \log(\hat{y})$, where $y$ is the true class label and $C$ denotes the total number of classes in the corpus.

*Contrastive Loss.* We hypothesise that similar credible news articles should be close to each other in the latent multimodal representation space, while unreliable news reports should be separated from them. Also, the previously published news articles could be referenced when making prediction for a recently released news report. We propose a constrative learning strategy. Given an article $x = \{h, o, I, t\}$, where $h$ is the headline, $o$ is the truncated body text, $I$ is the image in the article, and $t$ is the publication timestamp, we choose $k$ positive samples $s_{pos} \in S_{pos}$ that satisfy the following conditions: (i) published no later than $x$, $s_{pos}^{(t)} \leq x^{(t)}$; (ii) has the same credibility label as the target document; and (iii) among the document subset satisfying the aforementioned two conditions, choose the documents with their headlines among the top-$k$ most similar ones compared to that of $x$, calculated as the cosine similarity between the BERT encodings of $s_{pos}^{(h)}$ and $x^{(h)}$.

We do the same to select the top $k$ similar negative sample, $s_{neg} \in S_{neg}$. Except that this time, we select the top $k$ most similar articles with their credibility label opposite from that of the target article.

To speed up the selection of positive or negative samples, we use a memory bank to store the representations of training instances, donated as $M = \{M_1, M_2, ..., M_D\}$, where $D$ is the training set size. The initial values of the memory slots are all zeros. After the $i$-th iteration, the recomputed representation $x_d^i$ of a document $d$ overwrites the previous memory value $M_d^{i-1}$. Essentially, when fetching the top $k$ positive or negative training samples, we compare the representation of a document $d$ calculated in the $i$-th iteration with the representations of documents calculated in the $(i-1)$-th iteration.

To encourage similar credible news articles (or similar unreliable news) to be close, and news articles with similar headlines but with opposite class labels to be far away in the multimodal embedding space, we define the constrastive loss as:

$$\mathcal{L}_s = -\frac{1}{2k}\Big(\sum_{s_{pos} \in S_{pos}^x} \log \frac{\exp(\cos(x, s_{pos}))}{\sum_{s_{neg} \in S_{neg}^x} \exp(\cos(x, s_{neg}))}\Big) \quad (3)$$

where $S_{pos}^x$ denotes the set of positive training samples for document $x$, and similarly for $S_{neg}^x$, $\cos(\cdot)$ is the cosine similarity. To scale the contrastive loss into $[0, 1]$, we use the coefficient $\frac{1}{2k}$. $f(a, b)$ is a function calculating the difference between two input vectors, defined as $f(a, b) = 1 - \cos(a, b)$.

*Final Loss.* The final loss function is defined as a summation of the cross-entropy loss and the constrastive loss: $\mathcal{L} = (1 - \alpha)\mathcal{L}_c + \alpha\mathcal{L}_s$, where $\alpha$ is the weight controlling the contribution of each loss term.

## 4 EXPERIMENTS
### 4.1 Experimental Setup

*Dataset.* We conduct experiments on the COVID-19 related dataset, ReCOVery [28]. It contains a total of 2,029 news articles on coronavirus, published from January to May 2020. The articles are labeled as reliable or unreliable according to the source credibility. This repository provides multimodal information of news articles on coronavirus, including textual, visual, temporal, and network information. We clean the data and remove posts that contain unavailable image links. The final dataset contains 1,859 news articles in which 1,297 are labelled as '*Reliable*' and 562 as '*Unreliable*'.

*Models for Comparison.* In all our experiments, we follow the approach described in Sec. 3.2, to represent each image as a sequence of $m$ image patch vectors.[3] We conduct experiments using the following models, including: **CNN** [9], which is applied on the word sequence and the image patch vector sequence of a news article separately, and the results are concatenated and fed to a softmax layer for unreliable news detection; **LSTM** [10], which is also applied on the word sequence and image patch vector sequence separately, with the encoded results concatenated and fed to a softmax layer for classification; **SAFE** [28], the state-of-the-art model on the ReCOVery dataset, by first representing an image by a text description using a pre-trained image captioning model, and then concatenating image description with news text to train a text-based classifier for unreliable news detection; **BTIC**, our proposed **B**ERT-based **T**ext and **I**mage multimodal model with **C**ontrasting learning (§3).

In addition, we also consider the following variants of our proposed model: **BT**, BTIC using the text feature of $e^w$ to classify the news without the image information and contrastive loss. **BTI**, BTIC trained with the multimodal feature representation $e$, but without contrastive learning. **BTICr**, BTIC trained with the multimodal feature representation $e$, but using randomly selected negative samples to construct the contrastive loss.

*Parameter Setup.* In the text encoder, we set the text sequence length, $n = 256$, and the word embedding dimension, $d = 768$. In the image encoder, an input image is segmented into $m = 64(8 \times 8)$ patches, represented by a vector of $d^I = 2048$ dimensions, and then mapped into a $d = 768$ dimensional space. For contrastive learning, we set the sample set size $k$ to 5, set the weight $\alpha$ empirically to 0.2 in the final loss function. In optimisation, the learning rate is set to $10^{-6}$, and the number of epochs is 100.

### 4.2 Results

We repeat each experiment for 5 times, and report the averaged performance here (with p value less than 0.05 by Student's t-test.).

*Randomly shuffled data.* We first report the results in Table 1 on randomly shuffled data, following the setup in [28] that 80% of the data are used as the training set and the remaining as the test set. It can be observed that using the image patch vector sequence to represent an image is more effective than using an image description as both CNN and LSTM outperform SAFE significantly especially on the '*Unreliable*' category. Our proposed BTIC further improves upon CNN and LSTM by 2.8-3.8% in macro F1. Among the variants of BTIC, setting the loss weight value of $\alpha$ to 0.1 or 0.2 did not give any significant performance difference. But choosing negative samples randomly for the construction of the contrastive loss (BTIC$_{r0.2}$) hurts the performance on the '*Unreliable*' category with over 4% drop in recall. This shows that our proposed negative sampling method built on $k$-nearest neighbour search is more effective in dealing with the task of multimodal unreliable news detection.

---
[3]We have also experimented with representing an image by the vector generated from a pre-trained image encoder such as ResNet-50 or VGG, as typically done in traditional multimodal learning. But the results obtained are consistently worse than the image patch encoding method used here.

| Method | Mac. $F_1$ | Reliable news | | | Unreliable news | | |
|---|---|---|---|---|---|---|---|
| | | Pre. | Rec. | $F_1$ | Pre. | Rec. | $F_1$ |
| CNN | 0.736 | 0.813 | 0.928 | 0.866 | 0.766 | 0.508 | 0.606 |
| LSTM | 0.750 | 0.823 | 0.931 | 0.873 | 0.782 | 0.535 | 0.627 |
| SAFE | 0.633 | 0.759 | 0.942 | 0.840 | 0.702 | 0.308 | 0.425 |
| BTIC$_{0.1}$ | **0.778** | 0.836 | 0.938 | **0.884** | 0.804 | 0.577 | 0.671 |
| BTIC$_{0.2}$ | 0.774 | **0.843** | 0.912 | 0.876 | 0.755 | **0.609** | **0.672** |
| BTICr$_{0.2}$ | 0.770 | 0.832 | 0.934 | 0.880 | 0.791 | 0.566 | 0.659 |

**Table 1: Comparison with existing approaches on randomly shuffled data. BTIC$_\alpha$ denotes the results obtained with a specific weight value $\alpha$ set in the final loss function.**

*Chronologically ordered data.* We argue that it is more reasonable to split the data into the training and test sets in chronological order since future news articles should not be used to train a model to predict the credibility of historical news articles. We thus conduct another set of experiments on the dataset split by their publication timestamps with a ratio of 8:2. We further set aside 15% data from the training set as our validation set.

| Method | Mac. $F_1$ | Reliable news | | | Unreliable news | | |
|---|---|---|---|---|---|---|---|
| | | Pre. | Rec. | $F_1$ | Pre. | Rec. | $F_1$ |
| CNN | 0.666 | 0.807 | **0.834** | 0.809 | **0.653** | 0.512 | 0.523 |
| LSTM | 0.716 | 0.865 | 0.746 | 0.800 | 0.559 | 0.731 | 0.632 |
| SAFE | 0.556 | 0.729 | 0.929 | 0.817 | 0.566 | 0.203 | 0.295 |
| BTIC$_{0.1}$ | 0.728 | 0.861 | 0.780 | **0.818** | 0.586 | 0.708 | 0.639 |
| BTIC$_{0.2}$ | **0.735** | **0.876** | 0.763 | 0.815 | 0.580 | 0.750 | **0.654** |
| BTICr$_{0.2}$ | 0.690 | 0.870 | 0.692 | 0.767 | 0.522 | 0.756 | 0.614 |
| BT | 0.702 | 0.836 | 0.781 | 0.807 | 0.560 | 0.644 | 0.598 |
| BTI | 0.730 | 0.868 | 0.769 | 0.814 | 0.581 | 0.729 | 0.645 |

**Table 2: Ablation study on the chronologically ordered data. BTIC$_{0.1}$ and BTIC$_{0.2}$ denote the results obtained by setting $\alpha$ in the final loss function to 0.1 and 0.2, respectively.**

Table 2 show the results of all models including those of ablation study on our model. It can be observed that the performance drops consistently across all models compared to that on the randomly shuffled data. BTIC outperforms all the baselines. Among the BTIC variants, using text modality only (BT) gives the worst performance. Adding image information (BTI) boosts F1 on the '*Unreliable*' category significantly by nearly 5%. With contrastive learning and when the weight of the contrastive loss is set to 0.2 (BTIC$_{0.2}$), we observe a performance improvement over BTI with a significant increase of over 2% in recall in the '*Unreliable*' category. Simply using randomly selected negative samples (BTICr$_{0.2}$) lower the performance by 4.5% on macro F1 compared to BTIC.

Figure 1 shows the multimodal representations of samples in the test set learned by BTI (i.e., without contrastive learning) and BTIC, respectively. While points are scattered more widely in the embedding space in BTI with some points from different classes mixed with each other, they tend to be grouped more closely in two clusters in BTIC, showing the effectiveness of contrastive learning.

*Case study.* To illustrate the attention weights learned from the interaction layer of the transformer between text and image, we show in Figure 3 an example article and its accompanied image. The article has been classified correctly by our model as '*unreliable*'. In this case, "Anthony Fauci" receives higher attentions associated with the "*reliable*" class, both in the text modality and the image modality. However, the viewpoint expressed by Fauci in the interview, such as "shutdowns damage America", which cannot find the supporting evidence in the memory bank according to our contrastive loss, is categorised as "*unreliable*" as evidenced by the higher class-associated attention weight of the word 'damage'.

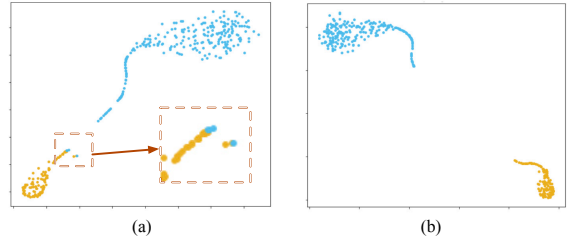

**Figure 2: The visualisation of learned embeddings by t-SNE: (a) The representation learned by BTI (without contrastive loss). (b) The representation learned by BTIC (with contrastive loss).**

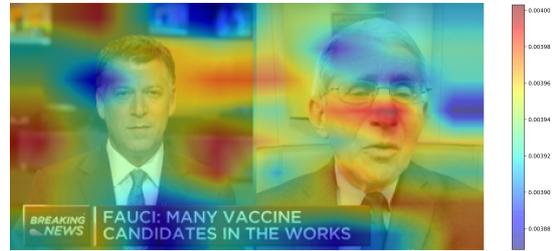

[CLS] …National Institute of Allergy and Infectious Diseases chief Dr. Anthony Fauci now says that "prolonged shutdowns" will damage America if it's not reopened soon…

**Figure 3: The heat map of attention weights from the interaction layer of the transformer between text and image (resized by bilinear interpolation). In text, words highlighted in blue have higher attentions weights with the '*unreliable*' category, while those in red are more closely related to the '*reliable*' category.**

## 5 CONCLUSIONS

In this work, we have proposed a novel framework for unreliable news detection, in which both text and visual information is fed into a BERT-based multimodal model to generate multimodal features that better encode the interactions between text and image. We have further incorporated contrastive learning to better learn multimodal representations by making use of articles published in the past and reported similar events. Experiments on a COVID-19 related dataset collected from a number of media outlets show the effectiveness of our proposed model. Nonetheless, there are two major limitations in this study that could be addressed in future work: (1) the dataset only contains COVID-19 related news, hence the representations learned may not generalise well to other news domains; (2) the news labels were derived automatically from the source credibility, which may not reflect the actual veracity of news articles.

## ACKNOWLEDGMENTS

This work was funded by the UK Engineering and Physical Sciences Research Council (grant EP/V048597/1, EP/T017112/1). YH is supported by a Turing AI Fellowship funded by the UK Research and Innovation (EP/V020579/1).